# Image Recognition using Region Creep

Kieran Greer, Distributed Computing Systems, Belfast, UK.
http://distributedcomputingsystems.co.uk
Version 3.0

***Abstract*—** This paper describes a new type of auto-associative image classifier that uses a shallow architecture with a very quick learning phase. The image is parsed into smaller areas and each area is saved directly for a region, along with the related output category. When a new image is presented, a direct match with each region is made and the best matching areas returned. Each area stores a list of the categories it belongs to, where there is a one-to-many relation between the input region and the output categories. The image classification process sums the category lists to return a preferred category for the whole image. These areas can overlap with each other and when moving from a region to its neighbours, there is likely to be only small changes in the area image part. It would therefore be possible to guess what the best image area is for one region by cumulating the results of its neighbours. This associative feature is being called 'Region Creep' and the cumulated region can be compared with train cases instead, when a suitable match is not found. Rules can be included and state that: if one set of pixels are present, another set should either be removed or should also be present, where this is across the whole image. The memory problems with a traditional auto-associative network may be less with this version and tests on a set of hand-written numbers have produced state-of-the-art results.

*Keywords***:** image classifier, auto-associative memory, quick learning, region creep.

## 1 Introduction

This paper describes a new type of auto-associative image classifier. It has similarities with the work in earlier papers [7][8] and uses the same type of shallow architecture with a very quick learning phase. Similar to Deep Learning [10], the image is parsed into smaller areas for regions, but each area is then saved as is for that region and only stores a link to a list of related output categories. When a new image is presented, a direct match with each region is made and the best matching areas returned. The category list for each selected area can then be retrieved and summed, to return a preferred category for the whole image. These areas can overlap with each other, where the area size and amount of overlap can be anything and when moving from a region to its neighbours, there is likely to be only small changes in the area image part. It would therefore be possible to guess what the best image area is for one region by cumulating the results of its neighbours. This is in fact an associative feature of the classifier that can re-construct missing or noisy input by using an interpreted match instead of the direct match and it is being called 'Region Creep'. These processes work at a local level however, of the region in question and its neighbours. Auto-association also requires a global picture and in this case that is done by adding global rules. These rules work at the whole image level and basically state that if one set of pixels are present, another set should either be removed or should also be present. While the rules appear to be specific, most of the construction can be done automatically. No weight sets are stored in the classifier, but when it is used for recall, counts are made for all matching areas and these decide the final selections. More work is therefore done at the classification stage. The architecture is therefore different to a traditional neural network, which may be fully-connected and would store weight values in the model. The auto-association process can change the input image to match more with the stored versions. A second matching phase can then start with the image produced by the first phase, and so on.

The rest of this paper is organised as follows: section 2 gives some related work, while section 3 describes the new classifier in more detail. Section 4 describes some test results, while section 5 gives some conclusions to the work.

## 2 Related Work

An associative network learns relation between objects, or input and output pattern sets. An auto-associative network learns relations between the same pattern set [9]. Deep Learning [10][13][14] has managed to almost master image recognition, but Decision Trees [4] are not far behind. At the heart of Deep Learning and the original Cognitron or Neocognitron architectures [5], is the idea of learning an image in discrete parts. Each smaller part is an easier task and cells can then be pooled into more complex cells with neighbourhoods. The architecture of [10] ends up with a top two layers that form an undirected associative memory. Auto-associative networks can be recurrent or feedforward. Recurrent networks include the Hopfield or Boltzmann Machine, for example. Feedforward networks use hidden layers to reduce the object relations to a set of feature vectors. To avoid confusion, the input patterns should not overlap, when the feature vectors also become orthogonal and Hebbian learning can be used. Because the input patterns should be unique, they are restricted in how much information they can store. Also therefore, they are only able to retrieve 'known' images, but can do so when only part of the image is input and so they can deal with noisy input.

The self-organising version of [12] may be similar to this version, with relation to the region creep. Auto-associative networks have several uses. For example, one paper [17] uses a variation of [12] to complete partly occluded images, which is similar to noisy or missing input. Another paper [16] suggests that they can be used to process video, or image sequences, and the paper [1] uses a new Oscillatory type for image edge detection. Voice and signal recognition are other uses.

The paper [6] describes some other shallow architectures that may include convolutions, but that can be tricked. While it has not been thoroughly investigated, the algorithm of this paper might be tricked in a different way. This paper's algorithm includes a bit more direction to its processing, where very subtle changes might not be so effective. Associative memory is the ability to recognise relations between objects, not just the objects themselves and global rules can provide this capability. This is however, a well-accepted method, for example [11]. The paper [15] describes what might be a similar idea called 'equivalence constraints'. Their program is a semi-supervised clustering process that uses gaussian mixture models. The equivalence constraints are global associations between pairs of nodes that are determined automatically, but external to the category measurements. It therefore provides additional external information that is used to improve the clustering performance.

The image classifier is closely related to two earlier papers by the author[7][8]. The paper [8] introduces a classifier with a shallow hierarchy that converts the input into discrete bands (cells) and links each band with its output category. It can learn the input in a single pass and notes the orthogonal nature of the grid architecture. For an analogy with the new classifier, bands are maybe replaced by areas and variables by regions. If the regions are not linked, then they must be able to behave independently as well. A second paper [7] gives a first version for the algorithm of this paper, using only pixel relations. Treating each pixel as a cell requires it to have a weighted association with the other pixels, which in that paper spanned the whole image. There is no overlap with single pixels, but using larger areas gives the region some definition and that can make it both distinct and allow for overlap. Therefore, a related calculation can take place that can replace the weight values.

## 3 The Image Recognition Algorithm

The new algorithm is therefore a localised image recognition process with global rules. Currently, it is being described for binary input only, or pixels with only 2 values and it forms an auto-associative memory. When training the classifier, an image is read and parsed into small areas. These areas are added to each region exactly as they are and linked to the image category. If the area already exists for a region, then the category list for that area is updated with a new category. Larger regions would reduce the number of calculations, for example, but if there are multiple areas in a region, they may need to be transposed first. Different views as part of an overlay would also be possible, because the regions are not dependent on each other. So far, testing has only considered regions and areas of the same size. There can also be gaps between regions, defined by a hop size, again to reduce processing time, but it will also affect a region's influence.

Adding areas is a very quick process, but determining the global rules requires comparing all images, to find differences in them and so this is much slower. The test phase then involves matching each region directly with the input image and using region creep and global rules to make that selection process more accurate. The current algorithm probably works best with very distinct images (black and white) and with only a single feature to learn, hence the single number datasets in section 4.

### 3.1 Region Creep

When using the classifier, the first stage is to match each test region directly with the output. This can simply be a pixel-for-pixel match and the result is a set of saved areas with the best total. An area can be of any size, where some tests varied from 3x3 to 5x5, but the matching process can only work if the input is already stored in the classifier. If there is noise or something new, then a direct match will not be as accurate. A second stage therefore uses region creep. With this, all of the surrounding regions to the region under consideration are retrieved, where the theory is that they should be similar. All surrounding regions therefore return their selected areas and these can be used to create an interpreted count for each pixel in the current region, which is a cumulative count for each time a surrounding area has a pixel present. These counts can then order the current region pixels into decreasing importance, or be used as weights for selecting valid pixels in areas. Matching train regions that have a score less than some factor, say 50% of the region creep score can be ignored as invalid. Then for the rest, compare each direct match with each region creep area and increment a count when both have the same pixel value. Select the best train areas, if any, that meet the threshold criteria. Note that when the image is re-constructed, more than 1 region has influence over any particular pixel addition.

### 3.2 Global Rules

The region creep is a local event and so, for example, some key feature on the other side of the image cannot be considered. Fully-connected neural networks, for example, may code these global associations automatically. Therefore, to help with this, global rules can be generated automatically, by comparing two images and noting where they differ. Differences in images are likely to be lines and curves, but to save adding all of these pixels, the corners of the differences can be recognised and used instead. Corners are where the difference joins with the rest of the image and where it may start to change direction, for example. Basically, if one set of these pixels is present, then the second set should not be present. If one image is true, then the other image interpretation should be false. It is actually likely to be more complicated than this, where two sets of pixels can both be positive and nested coordinate sets may also have to be considered. For this paper, a rule is only considered if one set of its pixels is 100% present. Then if more than 1 rule matches, do a union of the present pixels, where all must be present; and do an intersection of the missing pixel sets and remove only the intersection from the image.

### 3.3 The Training Stage Algorithm

The following algorithm was used to train the classifier:

Parse each image into areas of a specified size and add them to a region for the whole image.

1.1. The top left coordinate of the region and area can associate them, where same size is most usual.

1.2. If the region already contains the pixel area, add the category to it.

1.3. If the pixel area is new, then add the area and the category to it.

Create global association rules by comparing all whole images with each other and noting where they differ, using the following rules:

1.4. Create a new image with the difference areas only and mark the corner pixels of the difference lines.

1.5. Retrieve the regions that map best to these corner pixels, sort into a top *X* number and create 2 sets of pixels from them.

1.5.1. Add the 2 sets of pixels to a related list, as a rule.

1.5.2. If one set of pixels is present in an image, the second set is either missing or also present.

1.6. The only problem is when images share pixel sets.

1.6.1. To help, when using the global rules, add matching rules together, where pixels present is the union and missing is the intersection.

1.6.2. When the largest first set of pixels is present therefore, you can remove the smallest second set of pixels.

### 3.4 The Testing Stage Algorithm

The following algorithm was then used to test the classifier:

1. Parse the test image into parts and match each train region with it. Save the best matching train areas for each region.
2. Retrieve all train regions in the same coordinate space as the test area and count the pixel similarity.

2.1. If any direct match area has a score better than *X*, say 50%, then it can be considered.

3. For each test area, calculate a cumulative score from its surrounding regions and return this 'region creep' score.
   3.1. If any direct match region has a score less than *X*, say 50%, of the region creep, then it can be removed.
4. Select the train regions that match best with each test region.
5. A best match may not be exactly the same as the test region and so will change it during re-construction.
   5.1. If the best match is not direct, then use the train area pixels, but remove any category associations.
6. Apply region creep and global rules to make the final region selections.
   6.1. Re-construct the image completely, from this selection.
7. This final image can be used as the starting point for another run, or the category groups can be retrieved and summed, to select the most likely category.
8. Stop when final image of step 7 is the same 2 runs in a row, or after *Y* iterations.

# 4 Implementation and Testing

A computer program has been written in the C# language to implement the classifier. It can read an image dataset, train the classifier with it and then ask the classifier to either categorise the test images or perform auto-association corrections. Two classification tests were run on handwritten number datasets and they were both classified to the state-of-the-art, but only on the same dataset. A present pixel in the image would be represented by the number 1 in the data file and an empty pixel by the number 0. Firstly, the classifier was fed all of the images as the train dataset. After storing these, it was presented with the same set of images and asked to classify each one, where the test results are shown next.

## 4.1 Chars74K Hand-Written Numbers Dataset

This test used the set of hand-written numbers [18], but only the numbers 1 to 9. There were approximately 55 examples of each number and the binary image was converted into a 32x32 black and white ascii image first. Tests started with 3x3 areas but in fact 5x5 produced just slightly better results. The test took less than 1 hour to run on a standard laptop, but the setup was automatic, apart from the area size and some matching parameters (matching equation, 50% region creep requirement). The resulting accuracy level is close to what the best classifiers produce, with 96% accuracy overall, with a minimum accuracy of 85% and maximum accuracy of 100% in any one category. As a comparison, an earlier image recognition attempt [7] only produced a 46% accuracy over the same dataset. The classifier did not work well with a different set of test images however and so it would not be useful for generalising over previously unseen images. This might be expected for an auto-associative classifier. While the Deep Learning methods are able to recognise the number sets with an error percentage of only 1-2% (1.25%) [10], this method also has the image repair functionality. The result is summarised in TABLE 1.

### 4.2 Semeion Handwritten Digits Dataset

This dataset [2][3] was selected because the data files are already in the correct binary format. The dataset contains 1593 handwritten digits from 0 to 9, converted into 16x16 black and white ascii images. The train and test phase took only a couple of hours, using region sizes of 3x3. The new image classifier performed very strongly and classified the images to 99.8% accuracy. The recent paper [3] quoted a similar success score of 99.2% using SVM and Genetic Algorithms, but with a much quicker process. Mid-90% is quoted in other papers as well.

TABLE 1. Summary of Classification Results.

| Dataset | Score | Typical Score |
|---|---|---|
| Chars74K | 96% | 99% |
| Semeion | 99.8% | 99.2% |

### 4.3 Tests with Region Creep

A second test was to use region creep, to determine if it could repair an image. The classifier was trained to recognise 3 letters only – 'I', 'O' and 'T', each represented by an 8x8 grid. These are shown in Figure 1 and each region area was 3x3 in size, resulting in a total of 36 regions. The test phase allowed 3 iterations to change the image, where the final image for one iteration would be used as the input for the next iteration. After being trained on these letters, the classifier was asked to classify, firstly the test image of Figure 2. This was correctly identified as the letter ‘I’, but the iterations were also able to reconstruct the whole image from this noisy input, as shown by the image results for the 3 iterations.

The classifier however was shown to be sensitive to where the missing pixels might be placed. A second test, shown in Figure 3a, had the same number of missing pixels, but the lower RHS present pixel was moved from position 7-7 in the Test image of Figure 2, to 7-8 in Figure 3a. This has detached the pixel sufficiently from the main body that the local region creep cannot reach it and so it reconstructs the lower RHS to be empty instead, leading to the letter ‘J’ as the final result. The categories ‘I’ and ‘T’ also get an equal score, after the letter ‘J’ reconstruction, shown in Figure 3b. While the letter ‘J’ looks like a cross between an ‘I’ and a ‘T’ this is unlikely to be happening. The current equation does however, give a bias to even a single pixel that can be joined through creep, as opposed to an empty area, where this would need to be equal to expect letters to be joined consistently.

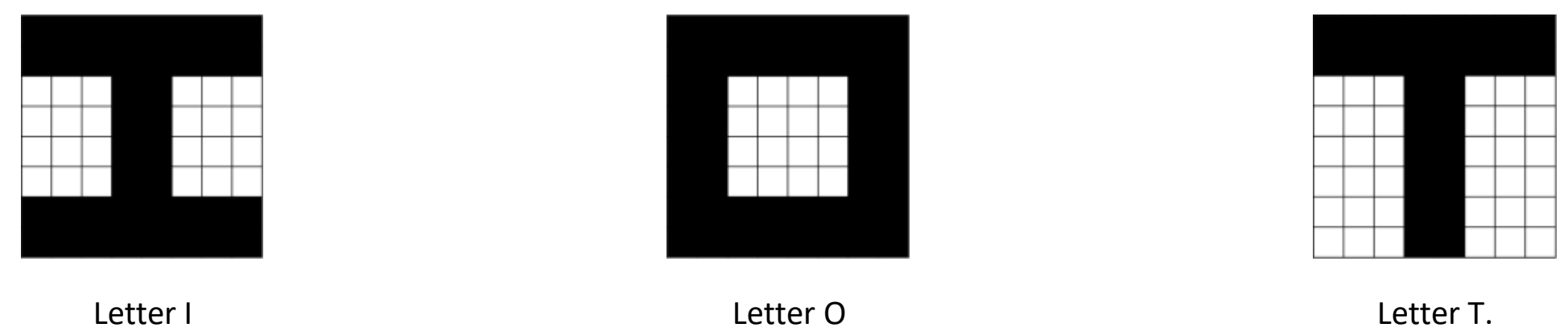

Figure 1. Letters for the auto-associative test, 8x8 grid.

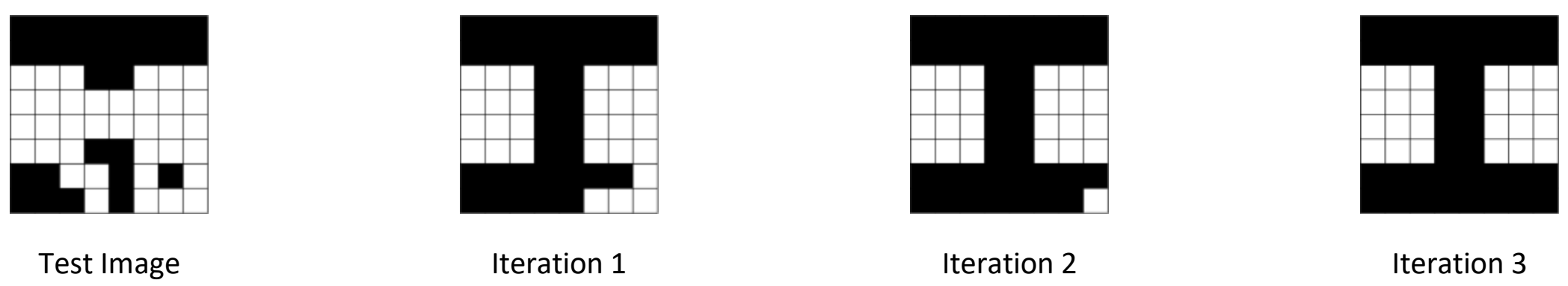

Figure 2. Associative memory reconstructs the letter I after 3 iterations.

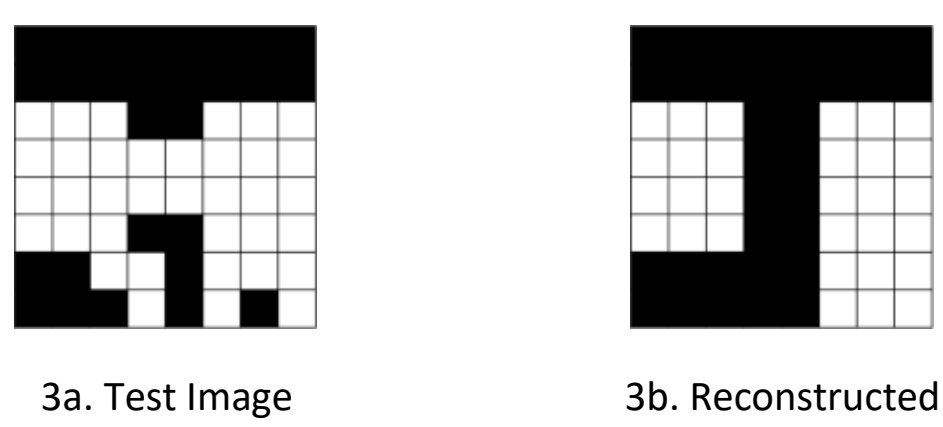

Figure 3. Associative memory reconstructs the previously unseen letter J.

## 5 Conclusions

This paper describes a new image classifier that is also an auto-associative memory, but the new design is not strictly a neural network. The shallow architecture is novel, where it maps pattern areas to category lists and the auto-association is through 'region creep'. Some of the differences are as follows:

1. The train stage stores image areas as they occur, without transposition through hidden layers, for example. They can then also be retrieved as is. Auto-association may not require hidden features, if it is to recognise the same image again.
2. Weight values are not saved in the model, but are calculated dynamically during recall, or the test stage.
3. The fully-connected nature of neural networks is converted into local 'region creep' aggregations with associated global rules.
4. A neural network classifier requires that the image and therefore feature sets are orthogonal, which is a memory restriction. This is also true for the new classifier, but the mapping from a region to a list of output categories adds another dimension that may increase memory capacity, in particular for the classification tasks.

While a reasonable amount of testing has been carried out, further work is necessary, in particular with the auto-association, and more complex images should also be tested. Test results show that the new algorithm can produce state-of-the-art results on the hand-written number datasets and with a minimal amount of setup or configuration.